\documentclass[11pt,reqno]{amsart}

\usepackage[utf8]{inputenc}   
\usepackage[T1]{fontenc}
\usepackage[english]{babel}

\usepackage[
  paper=letterpaper,
  textwidth=420pt,
  includehead,            
  headheight=8pt,
  headsep=14pt,
  textheight=584pt,       
  centering               
]{geometry}

\usepackage{amsmath,amsfonts,amsthm,amssymb}

\usepackage{graphicx}
\usepackage{epstopdf}  

\usepackage{caption}
\usepackage{subcaption}
\usepackage[colorlinks=true,linkcolor=blue,citecolor=blue]{hyperref}
\usepackage[capitalize,nameinlink]{cleveref}
\usepackage{algorithm}          
\usepackage{algpseudocode}      

\usepackage{enumitem}

\usepackage{amsthm}

\theoremstyle{plain}
\newtheorem{theorem}{Theorem}

\newtheorem{corollary}[theorem]{Corollary}

\theoremstyle{definition}

\theoremstyle{remark}
\newtheorem{remark}[theorem]{Remark}

\newcommand{\proj}{{\mathbf{P}}^\perp}

\newcommand{\cE}{\mathcal{E}}
\newcommand{\cF}{\mathcal{F}}

\newcommand{\bX}{\mathbf{X}}

\newcommand{\Sph}{\mathbb{S}}

\newcommand{\1}{{\rm 1}\kern-0.24em{\rm I}}

\newcommand{\R}{\mathbb{R}}
\newcommand{\p}{\mathbb{P}}

\newcommand{\T}{\mathbb{T}}

\newcommand{\pn}{\p_{\kern-0.25em n}}
\newcommand{\pnm}{\p_{\kern-0.25em n,m}}
\newcommand{\psubm}{\p_{\kern-0.25em m}}
\newcommand{\psubp}{\p_{\kern-0.25em p}}
\newcommand{\cfi}{\cF_{\kern-0.25em \infty}}

\newcommand{\argmax}{\mathop{\mathrm{arg\,max}}}

\newcommand{\ud}{\mathrm{d}}

\newlength{\minipagewidth}
\title{The Mean-Field Dynamics of Transformers}
\author{Philippe Rigollet}
\address{Massachusetts Institute of Technology}
\email{rigollet@mit.edu}

\date{\today}

\begin{document}

\begin{abstract}
We develop a mathematical framework that interprets Transformer attention as an interacting particle system and studies its continuum (mean-field) limits.  By idealizing attention on the sphere, we connect Transformer dynamics to Wasserstein gradient flows, synchronization models (Kuramoto), and mean-shift clustering.  Central to our results is a global clustering phenomenon whereby tokens cluster asymptotically after long metastable states where they are arranged into multiple clusters. We further analyze a tractable equiangular reduction to obtain exact clustering rates, show how commonly used normalization schemes alter contraction speeds, and identify a phase transition for long-context attention.  The results highlight both the mechanisms that drive representation collapse and the regimes that preserve expressive, multi-cluster structure in deep attention architectures.
\end{abstract}

\maketitle

\tableofcontents

\section{Introduction.}
Transformers, introduced by Vaswani et al.~\cite{vaswani2017attention}, have become the dominant architecture in modern machine learning, powering large language models and state-of-the-art systems across modalities. Their key novelty lies in the \emph{attention mechanism}, a data-dependent interaction between components of an input sequence. This mechanism allows each token---a vector representing a word, image patch, or general embedding---to update its representation by attending to all other tokens in the sequence.

From a mathematical standpoint, the attention mechanism can be viewed as defining a \emph{pairwise interaction} between tokens. Because neural networks act through an iterative composition of layers, their evolution may be interpreted as a discrete-time dynamical system, and in the continuous-time limit, as a nonlinear flow. This viewpoint underlies the theory of \emph{neural ordinary differential equations (neural ODEs)}~\cite{weinan2017proposal, haber2017stable, chen2018neural, li2018maximum, weinan2019mean, ruiz2023neural}, and provides a natural mathematical lens through which to analyze deep architectures. From this perspective, a Transformer may be seen as an \emph{interacting particle system}, in which each particle follows a velocity field depending on the empirical distribution of all others. This situates Transformers within the broad mathematical framework of mean-field dynamics.

The purpose of this paper is to present a mathematical framework that captures essential features of the Transformer architecture while remaining amenable to rigorous analysis. Our aims are twofold: to introduce the core ideas behind Transformers to a broad mathematical audience, and to highlight the connections between attention dynamics and areas of mathematics such as interacting particle systems, optimal transport, synchronization models, and gradient flows.

While our models are deliberately simplified, they preserve the essential structure of attention and layer normalization and thus remain directly relevant to practical Transformers. In particular, they exhibit the same qualitative \emph{clustering} observed in real networks. This simplified setting provides a concrete mathematical playground to study questions of long-time dynamics, metastability, and mean-field limits.

\smallskip

The paper is organized as follows. 
Section~\ref{sec:original_transformer} recalls the original Transformer
architecture, emphasizing the mathematical structure of the attention mechanism.
Section~\ref{sec:simplified_models} introduces simplified continuous-time models
that retain the essential features of attention and normalization, notably the
Self-Attention (SA) and Unnormalized Self-Attention (USA) flows, and discusses
their variational interpretation as Wasserstein gradient flows. 
Section~\ref{sec:clustering} describes the emergence of clustering in these
systems—empirically, qualitatively, and quantitatively—and establishes
connections with synchronization phenomena such as the Kuramoto model. 
Section~\ref{sec:metastability} develops the metastable picture of attention
dynamics, including slow motion between saddle points and saddle-to-saddle
transitions.
Section~\ref{sec:equiangular} introduces the equiangular model, which provides a
tractable one-dimensional reduction capturing clustering rates, the effect of
normalization, and the phase transition in long-context Transformers. 
We close this paper with Section~\ref{sec:noisy} on noisy Transformers.


\section{The original Transformer.} \label{sec:original_transformer}

The Transformer architecture, introduced by Vaswani et al.~\cite{vaswani2017attention}, is a modular neural network designed to process sequences of variable length. 
In contrast with recurrent and convolutional networks, whose structure enforces local or sequential dependencies, Transformers rely on a global interaction mechanism known as \emph{self-attention} or simply \emph{attention}. 
This mechanism allows every element of an input sequence—called a \emph{token}—to interact with every other token in a data-dependent way.

\paragraph{\bf Overall architecture.}
A standard Transformer is a stack of \emph{attention blocks}, each consisting of two main components: an \emph{attention layer} and a \emph{feed-forward MLP layer}. 
Both components are followed by normalization and residual connections. 
Schematically, if $\bX=(X_1, \ldots, X_n) \in\R^{d\times n}$ denotes the matrix of $n$ tokens of dimension $d$, a single layer performs the composition of the maps:
\[
\begin{aligned}
X_i &\longmapsto X_i + \mathrm{Attention}(\bX)_i,\\
X_i &\longmapsto X_i + \mathrm{MLP}(X_i)\,,
\end{aligned}
\]
for $i=1, \ldots, n$. In particular, all the tokens interact in the Attention layer but not in the MLP layer.
Normalization steps are inserted before or after each layer depending on the implementation; see Section~\ref{sec:norm} for some examples. 
These residual connections, together with layer normalization~\cite{ba2016layer, zhang2019root}, play a crucial stabilizing role, analogous to skip connections in residual networks~\cite{he2016identity, haber2017stable}.  
We will later see that such normalization can be idealized as a projection onto the unit sphere, a simplification that preserves essential dynamical features of trained models.

\paragraph{\bf Attention.}
The attention block computes for each token a weighted average of all other tokens.
Given learnable matrices $Q,K,V\in\R^{d\times d}$ (the \emph{query}, \emph{key}, and \emph{value} maps) and a temperature parameter $\beta>0$, the attention operator takes the form
\begin{equation}
    \label{eq:attention}
    \mathrm{Attention}(\bX)_i
    = \sum_{j=1}^n  \frac{\exp\!\big(\beta \langle QX_i, KX_j\rangle\big)}
    {\sum_{k=1}^n \exp\!\big(\beta \langle QX_i, KX_k\rangle\big)}  \, VX_j.
\end{equation} 
Equation~\eqref{eq:attention} can thus be interpreted as a nonlinear interaction rule in which each token updates as a weighted average of all others, with weights depending on their similarity in feature space.  
This operator will later serve as the starting point for the simplified continuous-time models studied in this paper.

In practice, the attention computation is distributed across several \emph{heads}, indexed by $h=1,\dots,H$, each with its own triplet $(Q_h,K_h,V_h)$.  
The resulting outputs are concatenated and linearly recombined.  
Multi-headed attention increases expressiveness and parallelism, and may be viewed as sampling several interaction kernels in parallel.

\paragraph{\bf MLP blocks.}

Following the attention block, a feed-forward network (usually a two-layer multilayer perceptron) acts independently on each token:
\[
\mathrm{MLP}(x) = W_2 \, \sigma(W_1 x + b_1) + b_2 ,
\]
where $\sigma:\R \to \R$ is a nonlinear activation function~\cite{hendrycks2016gaussian} 
applied to each entry of the vector $W_1 x + b_1$; a simple example to keep in mind is $\sigma(t) = \tanh(t)$ but many variations exist~\cite{kunc2024decadesactivationscomprehensivesurvey}.  
Together, the attention and MLP components define the local and global dynamics of the model.

\paragraph{\bf Encoder and decoder Transformers.}
Transformers were first developed for sequence-to-sequence tasks such as translation and therefore include an \emph{encoder–decoder} structure.  
The encoder transforms an input sequence into a latent representation through successive attention and MLP layers; the decoder then generates outputs by combining this representation with a causal (i.e., temporally restricted) version of attention.  
Variants such as BERT~\cite{devlin2019bert} use only the encoder, whereas GPT-type models employ the decoder alone with causal masking to enforce autoregressive behavior.  
Throughout this paper we focus on the encoder-type architecture and analyze the attention dynamics without causal masking, though we note that causal attention exhibits closely related phenomena~\cite{karagodin2024clustering}.

\paragraph{\bf Normalization.}
Normalization layers play a crucial role in stabilizing training and controlling the geometry of representations.  The two main variants are \emph{post-layer normalization} (which projects tokens back onto the sphere after applying attention) and \emph{pre-layer normalization} (which projects tokens onto the sphere before attention). Pre-layer normalization is used in leading models such as GPT~\cite{radford2019gpt} and LLaMA~\cite{touvron2023llama}. While this paper focuses on post-layer normalization, the mathematical results are not crucially dependent on that choice. In fact, we describe in Section~\ref{sec:norm} how several popular normalization rules can be understood in a common framework and all lead to some clustering, albeit at different speeds.

\paragraph{\bf Recent extensions.} Modern variants of the Transformer introduce additional components: hierarchical structures for vision tasks~\cite{liu2021swin}, sparse or long-range attention~\cite{beltagy2020longformer, child2019generating}, low-rank adaptation mechanisms~\cite{hu2022lora}, and mixture-of-experts~\cite{shazeer2017outrageously, boixadsera2025powerfinegrainedexpertsgranularity} architectures that route tokens through specialized subnetworks.  
These extensions have led to dramatic empirical gains but lie outside the scope of this paper.  
Our goal is to isolate and analyze the mathematical core of the architecture, the \emph{attention mechanism} itself, and to understand how it regulates token dynamics.

\section{Simplified models for attention dynamics.} \label{sec:simplified_models}

In this section we introduce idealized continuous-time models that capture the essential features of the self-attention mechanism. 
Our goal is not to reproduce every architectural component of a Transformer, but rather to isolate and formalize the mathematical structure that underlies its collective behavior. 
In particular, we seek to describe the evolution of tokens under attention as a system of interacting particles whose coupling depends on their pairwise similarities.

\subsection{From discrete layers to continuous time.}

As discussed in Section~\ref{sec:original_transformer}, a Transformer processes data through a sequence of layers, each performing an update of the form
\[
\bX_{k+1} = \bX_k + F_k(\bX_k),
\]
where $\bX_k \in \R^{n\times d}$ denotes the matrix of token embeddings at layer $k$, and $F$ encodes the combination of attention, feed-forward, and normalization operations.  
This recursive structure naturally suggests a discrete-time dynamical system.  
Following the analogy with residual neural networks~\cite{he2016identity, haber2017stable, chen2018neural, weinan2017proposal}, we interpret the layer index as a discretized time variable, and pass to the continuous-time limit
\[
\dot \bX_t = F_t(\bX_t).
\]
The resulting system may be viewed as a nonlinear flow on $(\R^d)^n$, where $n$ is the number of tokens.  
In this setting, the attention mechanism defines a nonlocal velocity field, coupling each particle to all others through a kernel that depends on their pairwise similarities.  
This perspective places Transformers within the theory of interacting particle systems and mean-field dynamics.
\subsection{Self-Attention (SA) dynamics.}

We now introduce a simplified continuous-time model that retains two essential components of the Transformer architecture: self-attention and layer normalization. 
For clarity we omit feed-forward layers and multi-headed structures, which may later be incorporated as additive or parallel terms without altering the core behavior.

Let $x_i(t)\in \Sph^{d-1}$ denote the position of the $i$-th token at time $t$, and let $\beta>0$ be an inverse-temperature parameter.
The \emph{self-attention (SA)} dynamics are given by
\begin{equation}\label{eq:SA}\tag{SA}
    \dot x_i(t) 
    = \proj_{x_i(t)}\!\left(
        \frac{1}{Z_{\beta,i}(t)} 
        \sum_{j=1}^n e^{\beta \langle x_i(t), x_j(t)\rangle}\, x_j(t)
      \right),
      \qquad 
      Z_{\beta,i}(t) = \sum_{k=1}^n e^{\beta \langle x_i(t), x_k(t)\rangle},
\end{equation}
where $\proj_x y = y - \langle x,y\rangle x$ denotes the orthogonal projection onto $T_x\Sph^{d-1}$.
The projection enforces the effect of layer normalization by keeping all tokens on the unit sphere.
The exponential weights represent attention scores, and the normalization ensures that each row of the attention matrix forms a probability vector.

Thus \eqref{eq:SA} describes $n$ particles on the sphere interacting through the kernel $K(x,y)=e^{\beta\langle x,y\rangle}$.
The interplay between this nonlocal interaction and the spherical geometry produces rich collective dynamics such as clustering and synchronization, which we study in later sections.

\medskip

A convenient variant omits the normalization step and projection, leading to the \emph{unnormalized self-attention (USA)} dynamics
\begin{equation}\label{eq:USA}\tag{USA}
    \dot x_i(t)
    = \proj_{x_i(t)}\!\left(\frac{1}{n}\sum_{j=1}^n e^{\beta \langle x_i(t), x_j(t)\rangle}\, x_j(t)\right),
\end{equation}
which is substantially easier to analyze and whose behavior often mirrors that of SA in practice.

\medskip

The empirical distribution of the tokens at time $t$ is
\[
\mu_t=\frac{1}{n}\sum_{i=1}^n \delta_{x_i(t)},
\]
and evolves according to the mean-field continuity equation
\begin{equation}\label{eq:continuity}
    \partial_t\mu_t + \nabla\!\cdot(\mu_t\, v_t[\mu_t]) = 0,
    \qquad
    v_t(x)=\proj_{x}\!\int e^{\beta\langle x,y\rangle} y\,\ud\mu_t(y).
\end{equation}
Because the velocity field depends nonlinearly on $\mu_t$, the equation is nonlinear and of McKean–Vlasov type.

This continuum formulation reveals an important structural distinction between the normalized and unnormalized models.  
For~\eqref{eq:USA}, the partial differential equation (PDE)~\eqref{eq:continuity} is the Wasserstein gradient flow~\cite{ambrosio2005gradient, CheNilRig25} of the interaction energy
\begin{equation}
\label{eq:E}
    \mathcal{E}_\beta(\mu)
    = \frac1{2\beta} \iint e^{\beta\langle x,y\rangle}\,\ud \mu(x)\ud \mu(y).
\end{equation}
For~\eqref{eq:SA}, it also corresponds to a gradient flow over the space of probability measures, albeit with respect to a different (Hessian) metric~\cite{geshkovski2025mathematical,Li21}. In the large-$\beta$ regime, these two dynamics recover familiar PDEs at leading order. Indeed the continuity equation associated with~\eqref{eq:SA} converges formally to a reverse heat equation. This anti-diffusive limit foreshadows the clustering behavior described in the next section. In contrast, for~\eqref{eq:USA} with appropriate time rescaling, it converges to a porous medium equation; see~\cite{bruno2025multiscale}.  

The~\eqref{eq:SA} and \eqref{eq:USA} dynamics are minimal but faithful abstractions of self-attention.  
They retain its essential nonlinear and nonlocal features while remaining amenable to analysis.  
As discussed in the next section, both exhibit clustering effects similar to those observed in trained Transformers, providing a tractable framework to study attention-driven representation dynamics.  In fact, when restricted to the circle, these dynamics reduce to variants of the Kuramoto model, the classical framework  to study \emph{synchronization} which corresponds precisely to the clustering behavior alluded to above.

\subsection{The Kuramoto connection.} \label{sec:circle}

When $d=2$, particles $x_i(t)\in\Sph^1$ are parametrized by angles $\theta_i(t)\in \T$, where $\T$ denotes the one-dimensional torus, i.e., the interval $[0,2\pi)$ with endpoints identified. 
The~\eqref{eq:USA} dynamics reduce to
\begin{equation}\label{eq:kuramoto}
    \dot{\theta}_i(t)
    = -\frac{1}{n}\sum_{j=1}^n 
      e^{\beta\cos(\theta_i(t)-\theta_j(t))}
      \sin(\theta_i(t)-\theta_j(t)).
\end{equation}
For $\beta=0$, this becomes
\[
\dot{\theta}_i(t)=
-\frac{1}{n}\sum_{j=1}^n 
\sin(\theta_i(t)-\theta_j(t))\,,
\]
the classical (homogeneous) \emph{Kuramoto model}~\cite{kuramoto1975self, acebron2005kuramoto}, 
originally introduced to study synchronization of coupled oscillators. 
It is known that for such dynamics, trajectories synchronize: for almost every initialization $(\theta_1(0), \ldots, \theta_n(0))\in\T^n$, one has $|\theta_i(t)-\theta_j(t)|\to0$ as $t\to\infty$~\cite{taylor2012there}. 

While the case $d=2$ in~\eqref{eq:kuramoto} is not of direct practical relevance to Transformers, it provides useful intuition and analytical tools for higher-dimensional attention dynamics. 
In particular, the temperature parameter $\beta$ modulates the system's complexity and governs the emergence of metastable states, as discussed in Section~\ref{sec:metastability}.

\section{Clustering in attention dynamics.} \label{sec:clustering}

\begin{figure}[h!]
    \centering
    \includegraphics[width=0.24\textwidth]{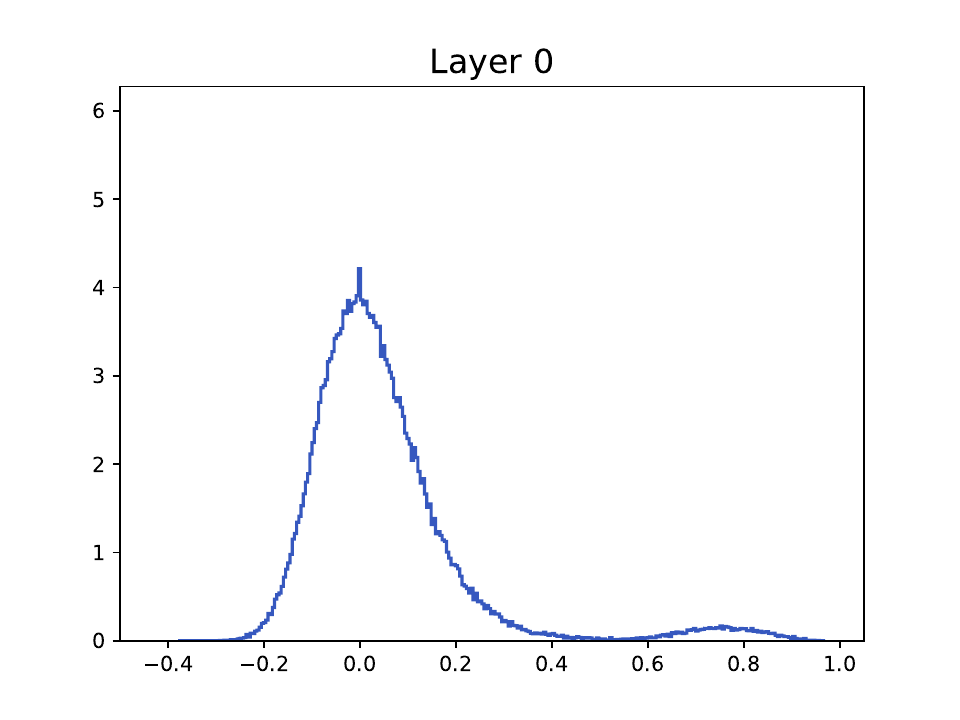}
        \includegraphics[width=0.24\textwidth]{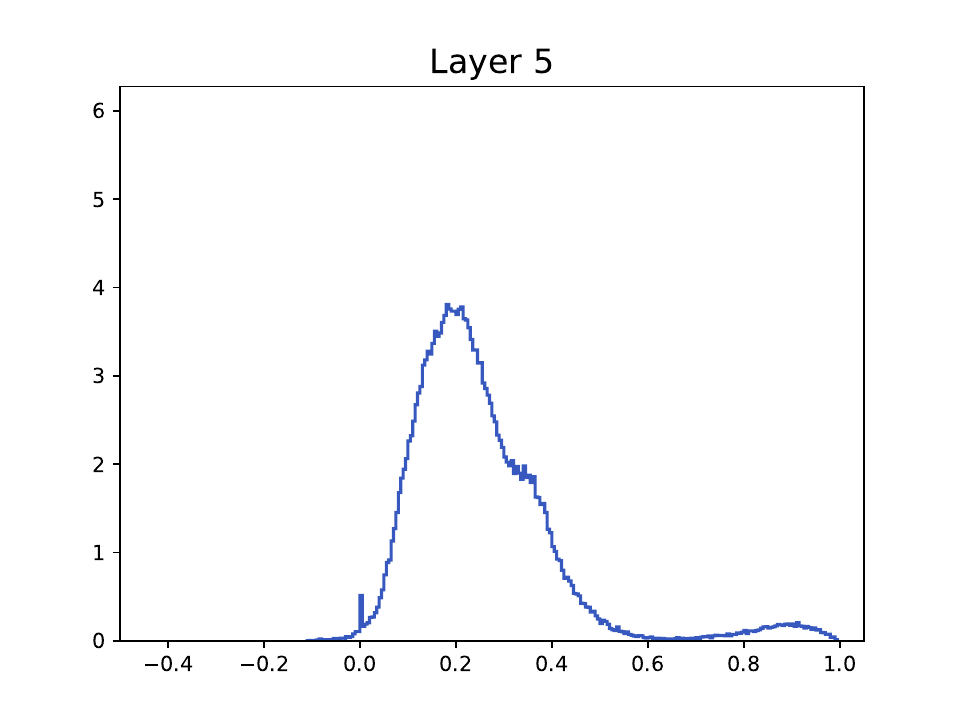}
    \includegraphics[width=0.24\textwidth]{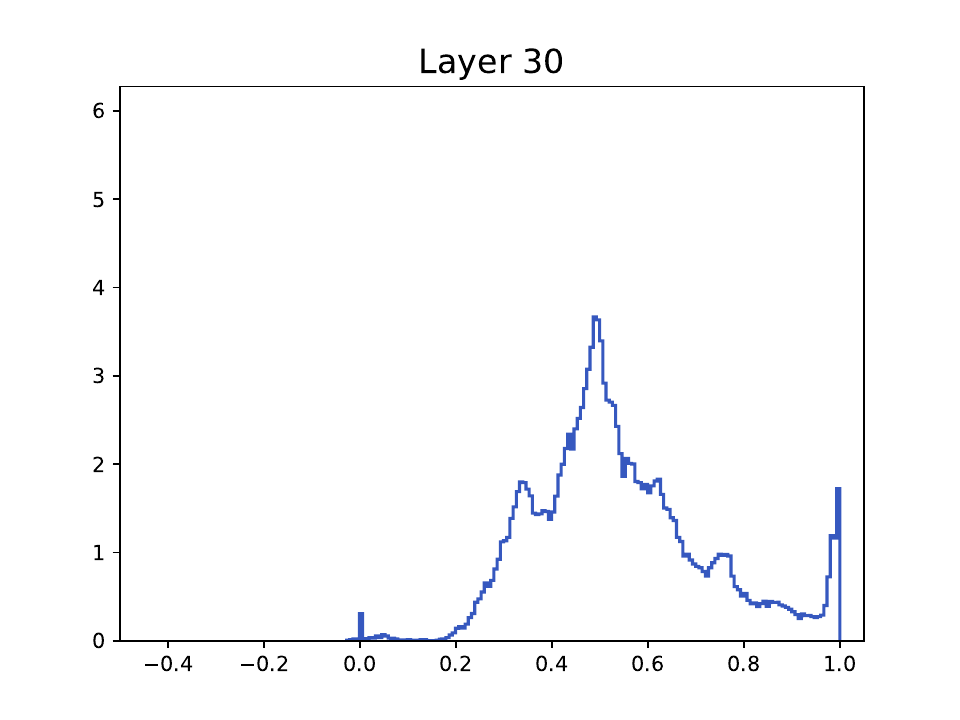}
    \includegraphics[width=0.24\textwidth]{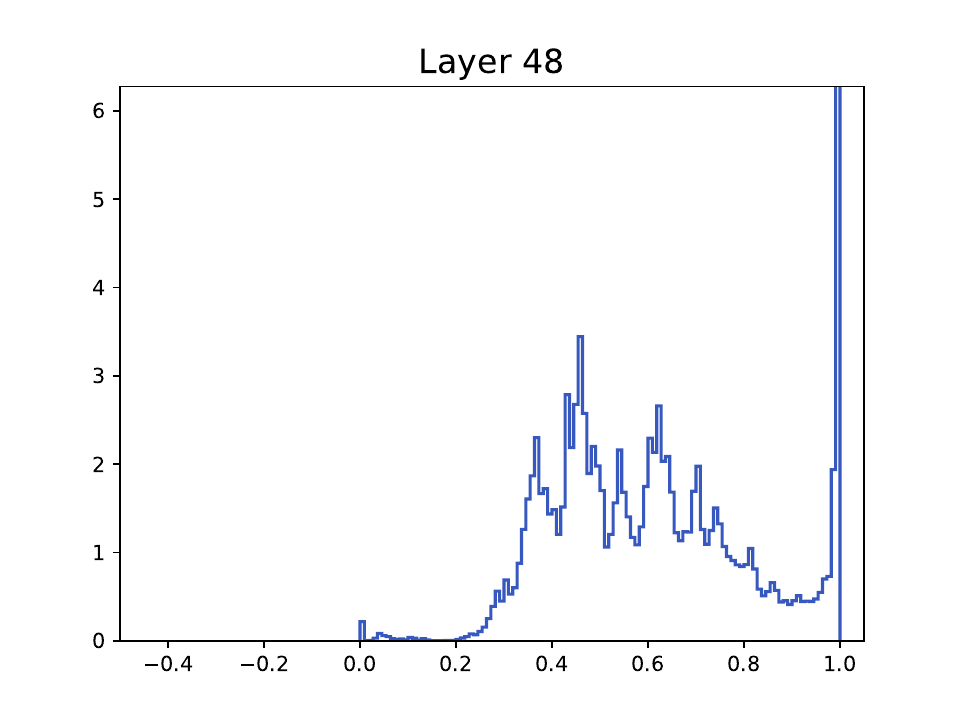}
    \caption{Histograms of pairwise inner products 
    $\{\langle x_i(t),x_j(t)\rangle\}_{i\neq j}$ 
    at layers $t=0$, $5$, $30$, and $48$ in the pre-trained {\sf ALBERT XLarge v2} model~\cite{lanalbert} available on Huggingface. 
    The progressive concentration of mass near~$1$ across layers illustrates the emergence of clustering in token embeddings.
    Figure reproduced from~\cite{geshkovski2025mathematical}.}
    \label{fig:albert}
\end{figure}

We now turn to one of the most striking properties of the dynamics~\eqref{eq:USA} and~\eqref{eq:SA}: the spontaneous emergence of \emph{clusters}. In fact, clustering can be observed in trained Transformer models: Figure~\ref{fig:albert} indicates token embeddings exhibit a progressive concentration of pairwise inner products near~$1$, revealing the gradual formation of clusters.

\subsection{A global clustering theorem.}

We now recall a general result guaranteeing convergence of the dynamics~\eqref{eq:SA} to a clustered state for all temperatures $\beta\ge 0$ and all ambient dimensions $d\ge3$.
The result originates in the study of synchronization on spheres, a line of work initiated by~\cite{markdahl2017almost} and recently refined in~\cite{criscitiello2024synchronization}.
It asserts that for a large class of smooth interaction laws depending only on pairwise inner products, the only asymptotically stable equilibria correspond to all particles being clustered in a single location (complete synchronization).

\begin{theorem}[\cite{markdahl2017almost,criscitiello2024synchronization,geshkovski2025mathematical}]\label{thm:clustering_finite}
The following holds for both~\eqref{eq:SA} and~\eqref{eq:USA} dynamics with $n\ge 2$ particles in dimension $d\ge 3$ and any $\beta\ge0$.
For almost every initial condition $(x_1(0),\dots,x_n(0))\in(\Sph^{d-1})^n$, the trajectories exist globally and converge to a clustered configuration:
\[
\lim_{t\to\infty} \|x_i(t) - x_j(t)\| = 0, \qquad \forall\, i,j\in[n].
\]
Equivalently, the empirical measure $\mu_t = \tfrac{1}{n}\sum_{i=1}^n \delta_{x_i(t)}$ converges weakly to a Dirac mass supported at some point $x_\infty\in\Sph^{d-1}$.
\end{theorem}
The result follows from the general theory of consensus dynamics on compact manifolds~\cite{markdahl2017almost}.  
Since the dynamics~\eqref{eq:SA} are a smooth gradient flow of an analytic energy functional,
the classical theorem of {\L}ojasiewicz~\cite{lojasiewicz1963propriete} guarantees that every trajectory converges to a stationary point.  
A detailed analysis of the critical set shows that, except for the clustered configuration, all other stationary points are saddles by exhibiting an escape direction. This result also holds for \emph{weighted tokens} that evolve according to the Wasserstein-Fisher-Rao gradient flow whereby particles evolve and are also dynamically reweighted~\cite{chen2025clustering}.
By the \emph{center-stable manifold theorem}~\cite[Thm.~III.7]{shub2013global}, the set of initial conditions whose trajectories converge to such saddles is contained in a countable union of lower-dimensional manifolds and thus has measure zero.  
Consequently, almost every initialization leads to convergence toward the clustered equilibrium.  
This argument, first established in~\cite{markdahl2017almost} and subsequently refined in~\cite{criscitiello2024synchronization}, provides a complete characterization of the long-time behavior of the~\eqref{eq:USA} dynamics; see~\cite{geshkovski2025mathematical} for an extension to the~\eqref{eq:SA} dynamics.

\begin{remark}
The case $d=2$ (the circle) is not covered by Theorem~\ref{thm:clustering_finite}, 
although it corresponds to the Kuramoto model ($\beta=0$) of Section~\ref{sec:circle} for which synchronization/clustering was proved in~\cite{taylor2012there}. Indeed, the proof of Theorem~\ref{thm:clustering_finite} fails when $d=2$.  
This gap was recently closed in~\cite{andrew25}, which proves clustering for all $\beta>-0.16$ in the case $d=2$ for both~\eqref{eq:USA} and~\eqref{eq:SA}.
\end{remark}

\subsection{Local rates of clustering.}
\label{sec:local_rates}
The finite-particle results above give almost-sure convergence but say nothing quantitative about rates.  In fact, using a Gr\"onwall argument, one obtains exponential convergence to a single cluster as soon as all tokens initially lie in a common open hemisphere.

\begin{theorem}[\cite{geshkovski2025mathematical}]\label{thm:cone-collapse}
Let $n\ge 1$, $\beta>0$ and $d\ge 2$.  Assume the initial tokens $(x_i(0))_{i=1}^n\subset\Sph^{d-1}$ satisfy
\[
\exists\; w\in\Sph^{d-1}\quad\text{such that}\quad \langle x_i(0),w\rangle>0\qquad\forall i\in[n],
\]
(i.e.\ all tokens lie in a common open hemisphere).  Let $(x_i(t))_{i=1}^n$ be the solution of either the~\eqref{eq:SA} or~\eqref{eq:USA} dynamics with this initialization.  Then there exist $x^*\in\Sph^{d-1}$ and positive constants $C,\lambda$ (depending on $n,\beta$ and the initialization) for which
\begin{equation}\label{eq:exp-collapse}
\|x_i(t)-x^*\|\le C e^{-\lambda t},\qquad\forall i\in[n],\ \forall t\ge 0.
\end{equation}
\end{theorem}

Since $n$ points in dimension $d\ge n$ must lie in the same hemisphere, we obtain the following corollary.
\begin{corollary}\label{cor:d-ge-n}
If the initial tokens are sampled i.i.d.\ uniformly on $\Sph^{d-1}$ and $d\ge n$, then they lie in some open hemisphere almost surely and Theorem~\ref{thm:cone-collapse} yields exponential convergence to a single cluster.
\end{corollary}

\subsection{Global rates of clustering in the mean-field limit.}

To obtain global rates of clustering, i.e., without a condition on the initialization, it is convenient to pass to the \emph{mean-field limit}, where the initial distribution of tokens $\mu_0$ that admits a density with respect to the uniform measure on the sphere.

For the Kuramoto model (the case $d=2$, $\beta=0$), Morales and Poyato~\cite{morales2022trend} established exponential rates of convergence of $\mu_t$ to a Dirac point mass.
In higher dimensions and for general $\beta>0$, the following theorem extends this result to self-attention dynamics.

\begin{theorem}[{\cite{chen2025quantitative}}]\label{thm:mfclust}
Let $d\ge 2$ and let $\mu_t$ evolve according to~\eqref{eq:continuity} from an initial measure $\mu_0$ with density $f_0\in L^2(\Sph^{d-1})$ satisfying
\[
R_0 := \Big|\int_{\Sph^{d-1}} x\, d\mu_0(x)\Big|^2 > 0.
\]
Then there exist constants $\beta_0, C_0, T_0>0$ depending on $\mu_0$ such that if $|\beta|<\beta_0$, there exists $x_\infty\in\Sph^{d-1}$ for which
\begin{equation*}\label{eq:mf_rate}
    W_2(\mu_t, \delta_{x_\infty}) \le C_0\, e^{-t/100}, \qquad t\ge T_0.
\end{equation*}
\end{theorem}

This result provides a \emph{quantitative} convergence rate for attention dynamics in the mean-field regime, complementing the qualitative clustering theorems for finite particles.  The argument extends that of~\cite{morales2022trend} to arbitrary dimensions but it is limited to small $\beta$. Indeed, an upper bound on $\beta$ is necessary: Example 2.6 in~\cite{chen2025quantitative} constructs an initialization on the circle showing that for large $\beta$ the mean-field dynamics may converge to multiple clusters. 
Such behavior suggests that large values of $\beta$ create an increasingly complex energy landscape in which several metastable states may appear as discussed in the next section.

\section{Metastability and the formation of multiple clusters.} \label{sec:metastability}

The previous section established that attention dynamics lead to clustering, either for finitely many particles or at the mean-field level.  
While these results characterize the asymptotic limit in which all particles collapse to a single cluster, empirical evidence (Figure~\ref{fig:albert}) and numerical simulations (white regions in Figure~\ref{fig: phase.diag.Id}) show that in practice, multiple clusters are typically observed.  
Such multi-cluster configurations are in fact desirable: when Transformers are viewed as measure-to-measure maps, they substantially enhance expressivity compared to the degenerate single-cluster limit.  
The early formation of these clusters has been rigorously analyzed in~\cite{bruno2025emergence}, which, in the mean-field limit, describes how small perturbations around the uniform initialization amplify into a structured, periodic arrangement whose number of clusters depends on the temperature parameter~$\beta$. A complementary analysis in~\cite{bruno2025multiscale} extends this picture to general parameter matrices and identifies multiple dynamical phases, including a slow phase associated with metastability, though it reduces to the models considered here when these matrices are the identity.  
Here, we focus our exposition on the long-time evolution of these clusters—specifically, on their slow coalescence and eventual collapse, a regime that reveals the metastable character of attention dynamics.

\subsection{Metastable dynamics and slow motion.}

The analysis in~\cite{geshkovski2024dynamic} establishes the existence of exponentially long-lived metastable states.  
Starting from a well-separated initial configuration consisting of $k$ sets of tokens that are close together relative to other sets, the tokens first collapse within each set, forming $k$ tight groups during the time interval $[0, T_1]$, where $T_1 \sim \beta$. 
These clusters then persist on an interval $[T_1, T_2]$ where $\log T_2 \sim \beta$ before successive merging events occur.  
Geometrically,  the flow of tokens remains near a manifold of $k$-cluster configurations, where the energy gradient is exponentially small and the motion of clusters is correspondingly slow. 

This metastable behavior can be understood from the energy perspective.  
A configuration with $k>1$ well-separated clusters corresponds to a nearly stationary point of $\cE_\beta$ where $\|\nabla \cE_\beta\|$ is exponentially small.  
Consequently, trajectories evolve extremely slowly in such regions of the energy landscape.  
This fits into the \emph{slow-motion} framework of Otto and Reznikoff~\cite{otto2007slow}, which states that for a gradient flow $\dot X=-\nabla \cE(X)$,  
if $\|\nabla \cE\|\le \delta\ll1$ on a manifold $\mathcal M$ and $\cE$ satisfies a Polyak--{\L}ojasiewicz-type inequality near $\mathcal M$,  
then trajectories remain trapped near $\mathcal M$ for times of order $\delta^{-1}$.  
Here $\delta\sim e^{-c\beta}$, explaining the exponentially long metastable time scale.

Beyond the metastable window $[T_1,T_2]$, the clusters slowly merge in a sequence of coarsening events,   each corresponding to a transition between nearby saddle points of the energy.  

\subsection{Saddle-to-saddle dynamics and the staircase profile.}

The energy functional $\mathcal{E}_\beta$ introduced in~\eqref{eq:E} admits a hierarchy of saddle points of increasing energy, connected by heteroclinic orbits that describe the gradual merging of clusters.  
After a suitable time rescaling, the energy evolves through long plateaus corresponding to metastable phases, separated by abrupt jumps each time two clusters coalesce.  
These saddle-to-saddle transitions become sharpest in the limit of the gradient flow of $\mathcal{E}_\beta$ when $\beta \to \infty$ and time is properly rescaled.  
In this regime, the configuration remains effectively frozen until it moves abruptly to the next saddle in the hierarchy by merging two clusters.

\begin{figure}[h!]
\centering
\includegraphics[scale=0.75]{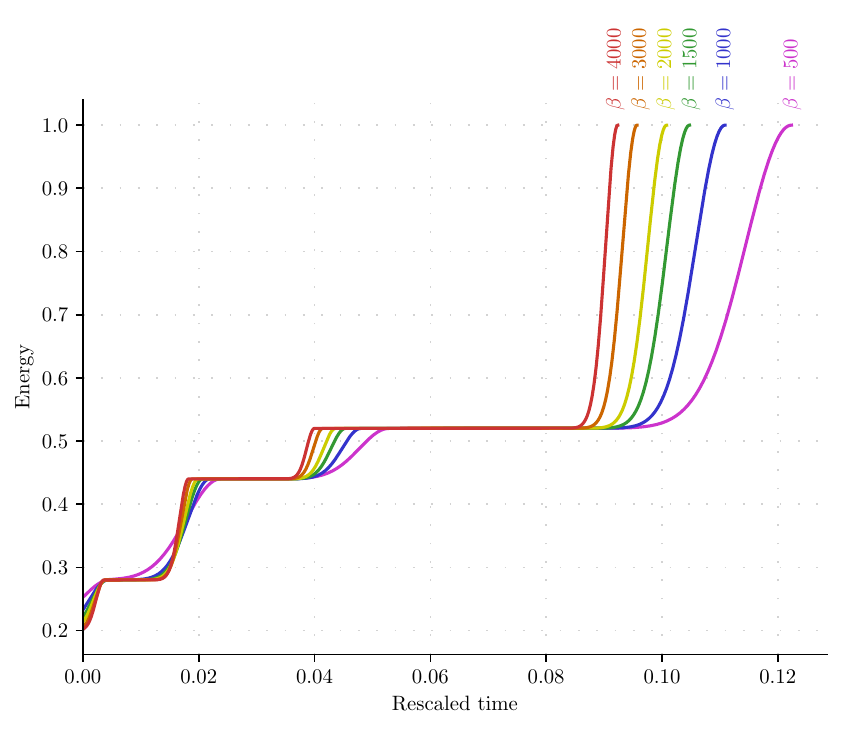}
\caption{
Energy $\mathcal{E}_\beta(t)$ along a metastable trajectory in the low-temperature limit $\beta\to\infty$.  
The energy remains constant over long plateaus corresponding to metastable multi-cluster configurations and increases sharply when clusters merge, forming a staircase profile.  
Each jump corresponds to a transition between successive saddle points of the energy landscape.  
Figure reproduced from~\cite{geshkovski2024dynamic}.}
\label{fig: staircase}
\end{figure}

Consider an initialization of the form
\begin{equation}
    \label{eq:init_clust}
    \mu_0=\sum_{j=1}^K \alpha_j \delta_{x_j(0)}, 
    \qquad 
    \alpha_j \ge 0,\quad \sum_{j=1}^K \alpha_j =1\,.
\end{equation}
This represents a configuration composed of $K$ clusters of respective masses $\alpha_1,\ldots,\alpha_K$ and captures, in particular, initializations located at a saddle point.  

Following~\cite{geshkovski2024dynamic}, Bruno, Pasqualotto, and Agazzi~\cite{bruno2025multiscale} provide a complete multiscale dynamical picture describing the evolution from such an initialization to the final single-cluster state in the limit $\beta \to \infty$.  
In particular, their analysis identifies a final \emph{pairing phase} governed by hardmax-like dynamics~\cite{hardmax}, during which the two closest clusters merge.  
This corresponds to a transition from one saddle of $\mathcal{E}_\beta$ to the next higher-energy saddle.

To build intuition, notice that as $\beta\to\infty$ the softmax velocity field in~\eqref{eq:SA} converges\footnote{Formally, the limiting rule is
\[
x_i(t)=\proj_{x_i(t)}\!\left(\argmax_{x_j(t)} \langle x_i(t), x_j(t)\rangle\right)
     = \proj_{x_i(t)} x_i(t)=0\,.
\]
Following~\cite{hardmax}, and to avoid this triviality, we forbid tokens from attending to themselves.}
to a hard $\argmax$ rule:
\[
\dot x_i(t)
= \proj_{x_i(t)}
  \Bigl(\argmax_{x_j(t) \neq x_i(t)} \langle x_i(t), x_j(t)\rangle\Bigr)\,.
\]
Because $\beta$ is large, the (unique) closest pair $(\bar{\imath},\bar{\jmath})$ in the sense that
\[
\langle x_{\bar{\imath}}(0),x_{\bar{\jmath}}(0)\rangle
= \max_{i\neq j}\langle x_i(0),x_j(0)\rangle,
\]
interacts on a timescale that is exponentially faster in $\beta$ than all other pairs.  
This leads to a deterministic merging event of that closest pair before any other interaction occurs.

\begin{theorem}[\cite{bruno2025multiscale}]
    \label{thm:agazzi_merge}
    Assume the initial datum is a discrete multi-cluster configuration of the form~\eqref{eq:init_clust}.
    Suppose $(\bar{\imath},\bar{\jmath})$ is the unique pair maximizing the inner product $\langle x_i(0),x_j(0)\rangle$ and rescale time by
    \[
        \ud t = e^{\beta(1-\langle x_{\bar{\imath}}(0),x_{\bar{\jmath}}(0)\rangle)}\, \ud s .
    \]
    Then, as $\beta \to \infty$, the trajectories $x_i(t), i=1, \ldots K$ converge, uniformly on any interval $[0,T_\varepsilon]$ on which 
    $\langle x_{\bar{\imath}}(s),x_{\bar{\jmath}}(s)\rangle \le 1-\varepsilon$,  
    to the solution of
    \[
    \dot y_k(s)=
    \begin{cases}
      \proj_{y_{\bar{\imath}}(s)}(y_{\bar{\jmath}}(s)), & k=\bar{\imath}, \\[3pt]
      \proj_{y_{\bar{\jmath}}(s)}(y_{\bar{\imath}}(s)), & k=\bar{\jmath}, \\[3pt]
      0, & \text{otherwise},
    \end{cases}
    \qquad
    y_k(0)=x_k(0).
    \]
    In particular, all clusters remain stationary except for the closest pair $(\bar{\imath},\bar{\jmath})$, which move along the unique geodesic connecting them and merge in finite rescaled time.
\end{theorem}

\subsection{Connection to Mean-Shift clustering.}
\label{sec:mean_shift}

The dynamics~\eqref{eq:SA} are closely related to a continuous-time analogue of the classical \emph{Mean-Shift} clustering algorithm.  
In its original form, the Mean-Shift algorithm defines clusters as basins of attraction of the modes of a kernel density estimator (KDE)~\cite{fukunaga1975estimation}.  
More precisely, given points $x_1(0), \ldots, x_n(0)\in \R^d$ drawn independently from a density $p$, and a kernel $K(\cdot)$ on $\R^d$ (typically Gaussian), recall that the KDE of $p$ is given by $\hat p =  K * \mu_0$, the convolution of $K$ with the empirical measure $\mu_0$ of the $x_i$; see, e.g.,~\cite[Chapter~1]{tsybakov2008nonparametric}.  
The following gradient-flow dynamics move each point toward the nearest mode of $K * \mu_0$:
\[
    \dot x_i(t) = \nabla \log (K * \mu_0)(x_i(t)), \qquad i=1,\ldots,n.
\]
This algorithm, along with suitable time discretizations, can be analyzed using tools from optimization and statistics.  
Notably,~\cite{arias2016estimation} establish consistency of the estimated gradient lines and show that fixed-KDE Mean-Shift recovers the modal structure of the underlying density under classical smoothness assumptions.

A modification proposed by Cheng~\cite{cheng1995mean}, often called \emph{blurring} Mean-Shift, recomputes the KDE at every iteration using the updated points.  
This leads to the mean-field dynamics
\[
    \dot x_i(t) = \nabla \log (K * \mu_t)(x_i(t)), \qquad 
    \mu_t = \frac{1}{n} \sum_{i=1}^n \delta_{x_i(t)}.
\]
When the points are constrained to lie on the sphere and $K$ is Gaussian, these dynamics coincide exactly with~\eqref{eq:SA}.  
Indeed, taking $K(x) = \exp\!\left(-\frac{\beta}{2}\|x\|^2\right)$ and interpreting $\nabla$ as the Riemannian gradient on the sphere, one obtains the velocity field
\[
    \nabla \log (K * \mu_t)(x_i(t))
    = \nabla \log \frac{1}{n} 
      \sum_{j=1}^n \exp\!\left(-\frac{\beta}{2}\|x_i(t) - x_j(t)\|^2\right).
\]
Using the identity $\|x_i(t)-x_j(t)\|^2 = 2 - 2\langle x_i(t), x_j(t)\rangle$ for points on the sphere, we recover precisely the vector field appearing on the right-hand side of~\eqref{eq:SA}.

This analogy suggests tokens cluster into $M$ groups, where $M$ is the number of modes of the KDE $\hat p$.  
Using Edgeworth expansions and the Kac–Rice formula, we  can show that $\mathbb{E}[M]$ is of order  $\sqrt{\beta \log \beta}$ as $n \to \infty$, in the regime $n^c \le \beta \le n^{2-c}$ (with arbitrarily small fixed $c$), when $p$ is a Gaussian density on the real line~\cite{geshkovski2025numbermodesgaussiankernel}.  
Numerical simulations indicate that this order of magnitude should also hold on the circle, at least up to logarithmic factors. Recall, however, that this multiple-cluster state is metastable and will eventually collapse to a single one according to the results of Section~\ref{sec:clustering}.

While this approach is currently limited to the one-dimensional case, it gives an approach to compute the number of metastable states in self-attention dynamics. An alternative approach to understand the first metastable state was proposed in~\cite{bruno2025emergence} by linearizing~\eqref{eq:continuity} around the uniform distribution. This different approach also points to a metastable state that contains $\Theta(\sqrt{\beta})$ clusters.

\section{Equiangular model.}
\label{sec:equiangular}
We now turn to an even simpler model that nonetheless captures the core mechanism behind clustering in attention dynamics.  
In this setting, the evolution of tokens reduces to a one-dimensional process, allowing for a much sharper analytical description of the clustering behavior. 

This equiangular model was initially introduced in~\cite{geshkovski2025mathematical} to study the evolution of (near-)orthogonal token initialization in attention dynamics and was later used in \cite{cowsik2024geometric} and \cite{giorlandino2025failuremodesdeeptransformers} to examine the role of large random weight matrices using formal heuristic calculations.

\subsection{Exact rates of clustering.}

Consider an equiangular initialization\footnote{Note that since $\rho_0\ge 0$ all points are initialized in the same hemisphere, as in Section~\ref{sec:local_rates}.}  where 
$$
\langle x_i(0),x_j(0)\rangle=\rho_0 \in [0,1] \quad \forall \   i\neq j.
$$
Because of symmetry, the configuration remains equiangular for all times under both~\eqref{eq:SA} and~\eqref{eq:USA} dynamics, 
and the entire system is characterized by the common correlation $\rho(t):=\langle x_i(t),x_j(t)\rangle$, for all $i\neq j$.
The evolution of $\rho$ is described by a simple ordinary differential equation (ODE):
\begin{align*}
\dot{\rho}(t)
&=\frac{2e^{\beta\rho(t)}(1-\rho(t))\big((n-1)\rho(t)+1\big)}
{e^\beta+(n-1)e^{\beta\rho(t)}}&\text{for~\eqref{eq:SA}} \\[0.3em]
\dot{\rho}(t)
&=\frac{2}{n}e^{\beta\rho(t)}(1-\rho(t))\big((n-1)\rho(t)+1\big),&\text{for~\eqref{eq:USA}}
\end{align*}
with initial condition $\rho(0)=\rho_0$.
Linearizing near the clustered state $\rho=1$ and setting $\varepsilon(t)=1-\rho(t)$, we obtain
\[
\dot\varepsilon(t)\simeq -2\varepsilon(t)
\quad\text{for \eqref{eq:SA},}\qquad
\dot\varepsilon(t)\simeq -2e^{\beta}\,\varepsilon(t)
\quad\text{for \eqref{eq:USA}.}
\]
Hence $1-\rho(t)\lesssim e^{-\lambda_\beta t}$ with an explicit exponential rate $\lambda_\beta>0$.  
This one-dimensional reduction already captures exponential convergence to complete clustering, in full agreement with Theorem~\ref{thm:cone-collapse}.  
Interestingly, the two models exhibit markedly different behaviors: for~\eqref{eq:USA} the rate $\lambda_\beta=2e^{\beta}$ grows exponentially with~$\beta$, while for~\eqref{eq:SA} it remains constant.  
This distinction is favorable to~\eqref{eq:SA}, as excessively fast contraction tends to accelerate representation collapse in deep Transformer architectures; see Section~\ref{sec:norm}.

\begin{figure}[h!]
    \centering
        \includegraphics[width=0.3\textwidth]{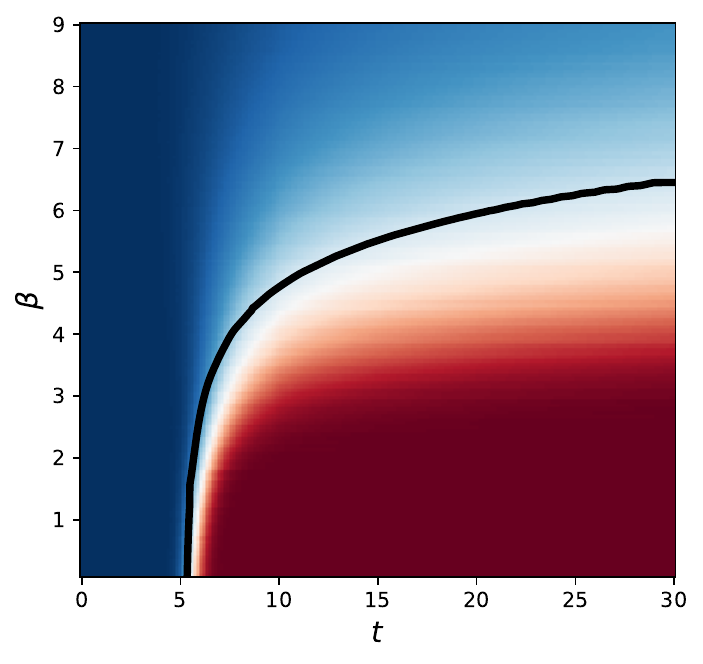}
    \includegraphics[width=0.3\textwidth]{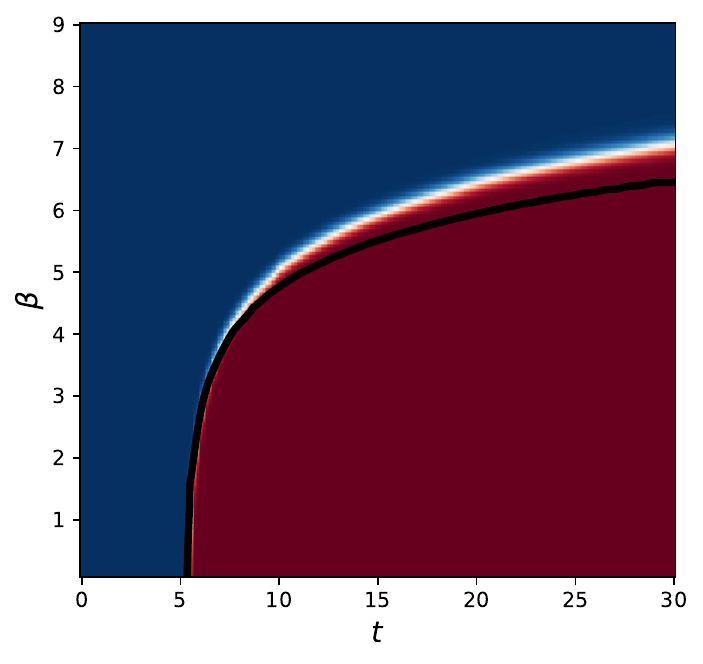}
    \includegraphics[width=0.342\textwidth]{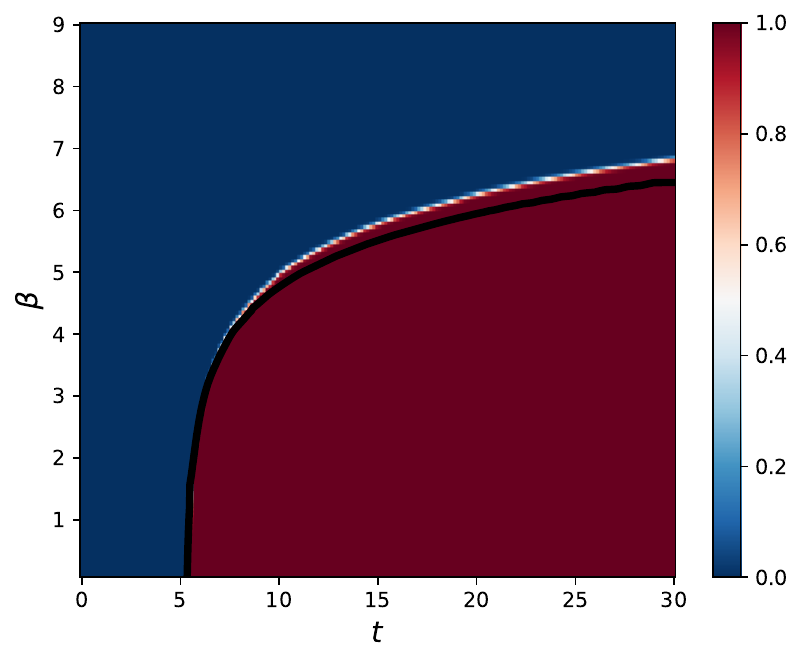}
    \caption{Phase transition diagrams for randomly initialized particles following~\eqref{eq:SA}.  
Each panel shows the probability that $\langle x_i(t),x_j(t)\rangle \ge 0.999$ on a fine grid of values for the pair $(t,\beta)$ for $n=32$ tokens.  
As the dimension $d$ increases from $d=8$ (left), to $d=128$ (middle) to $d=1,024$ (right), the transition curve sharpens and aligns with the theoretical prediction from the equiangular model.  
Reproduced from~\cite{geshkovski2025mathematical}.}
    \label{fig: phase.diag.Id}
\end{figure}

As $d\to\infty$, random points on the sphere become almost orthogonal by concentration of measure: $\langle x_i(0),x_j(0)\rangle = \rho_0 = 0$ for all $i\neq j$ and the behavior predicted by the equiangular model is clearly visible in numerical experiments in Figure~\ref{fig: phase.diag.Id}.

\subsection{The impact of normalization.}
\label{sec:norm}

Normalization layers are a defining component of modern Transformers.  
The original formulation of attention as an interacting particle system without normalization was introduced in~\cite{sander2022sinkformers}.  
In that setting tokens move freely in $\R^d$ and typically either diverge or collapse to the origin, and clustering results require additional assumptions and can be elicited on rescaled tokens~\cite{geshkovski2024emergence, castin2025unified}.  
By contrast, the presence of layer normalization (LN) as in real architectures keeps representations on a controlled scale and produces crisp clustering behavior.

Both~\eqref{eq:SA} and~\eqref{eq:USA} above employ Post-LN where tokens are projected onto the sphere \emph{after} the attention layer. This is to be contrasted with Pre-LN where tokens evolve freely in $\R^d$ but the attention scheme is applied to normalized tokens. In this case, the~\eqref{eq:USA} dynamics become
$$
    \dot x_i(t) = \frac{1}{n} \sum_{j=1}^n e^{\beta \langle \frac{x_i(t)}{\|x_i(t)\|}, \frac{x_j(t)}{\|x_j(t)\|}\rangle}\, \frac{x_j(t)}{\|x_j(t)\|},
$$
and the~\eqref{eq:SA} dynamics are modified similarly with appropriate normalization. Many other variations exist: for example, Peri-LN employs normalized tokens in attention like Pre-LN but it also normalizes the output of the attention itself, resulting in $x_i(t) \in \R^d$ that is subject to a unit-norm velocity field.

To understand the effect of normalization, write each token as $x_i(t)=r_i(t)\,\theta_i(t)$ with $\theta_i(t)\in\Sph^{d-1}$ and $r_i(t)>0$.   All decoding stages depend only on directions $\theta_i$, so it is natural to track their evolution.   A key observation is that \emph{every normalization rule} induces the same underlying attention vector
\[
A_i(\Theta)
= 
\frac{\sum_{j=1}^n e^{\beta\langle\theta_i,\theta_j\rangle}\,\theta_j}
     {\sum_{k=1}^n e^{\beta\langle\theta_i,\theta_k\rangle}},
\]
where $\Theta(t)=(\theta_1(t), \ldots, \theta_n(t))$.  
In all cases the directions satisfy the normalized-attention ODE
\begin{equation}\label{eq:NA}
\dot\theta_i(t)
= \frac{1}{s_i(t)}\proj_{\theta_i(t)}\,A_i(\Theta(t))\,,
\end{equation}
while the magnitudes $r_i(t)$ satisfy a rule-dependent radial equation.  
Crucially, each normalization scheme induces a corresponding \emph{speed regulation factor} $s_i(t)$ on token $i$. For example, for Post-LN, taking $s_i(t)=1$ recovers~\eqref{eq:SA} while it can be shown that Pre-LN yields $s_i(t)=r_i(t)$ so that the directions of tokens with large magnitude are effectively slowed down.
Other variants include Peri-LN~\cite{kim2025peri}, Mix-LN~\cite{li2024mixln}, nGPT~\cite{loshchilov2024ngpt}, and sqrt-scaling~\cite{Nocisqrt}. All fit into this framework with different choices of $s_i(t)$.

Using this unified framework,~\cite{karagodin2025normalization} show that attention dynamics collapse to a single cluster in long time even in the presence of speed regulation factors. Moreover, studying these dynamics under the equiangular model yields a sharper image that reveals important distinctions between the various LN schemes.

\begin{figure}[ht]
   
        \centering
        \includegraphics[width=0.5\textwidth]{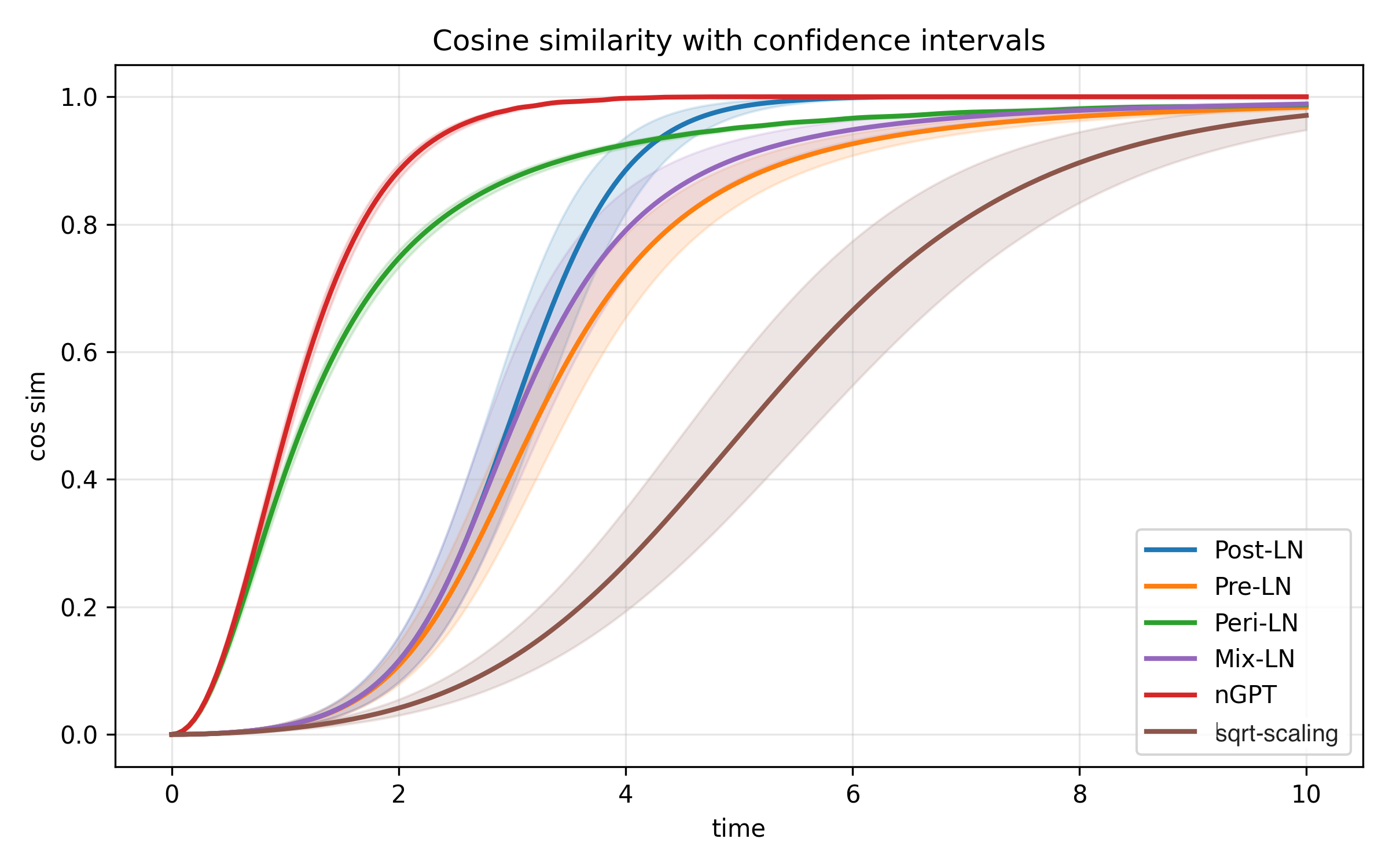}
    \caption{Evolution of average cosine similarity $\rho(t)$ with 90\% confidence interval on a Transformer with random weights, $d = 512, \beta = \sqrt{d}, d>n$ and random initial tokens. Figure reproduced from~\cite{karagodin2025normalization}.}
    \label{fig:norm}
\end{figure}

Under an equiangular initialization, we get $\langle \theta_i(t),\theta_j(t)\rangle=\rho(t)$ for all $i\ne j$ and $r_i(t)=r(t)$ for all tokens $i$. The coupled ODEs governing $\rho(t)$ and $r(t)$  can be written explicitly for each scheme and solved explicitly as $t \to \infty$ using a simple linearization argument. It yields $1-\rho(t) \sim e^{-2t}$ for Post-LN while $1-\rho(t)\sim 1/t^2$ for Pre-LN.
This marked difference in the rate of contraction to a single cluster---exponential vs. polynomial---confirms the practical wisdom that Pre-LN makes better use of depth by delaying contraction and hence avoiding representation collapse; see Figure~\ref{fig:norm}.

The equiangular model appears to be predictive of the rate of clustering despite its apparent simplicity. In fact, the linearization argument may be applied beyond the equiangular setup and similar conclusions can be made in these more general dynamics~\cite[Theorem~4.3]{karagodin2025normalization}. 

\subsection{Long context Transformers.}
The equiangular model also sheds light on the behavior of \emph{long-context} Transformers, where $n$ is large and attention scores tend to flatten.  
In such regimes, the softmax denominator grows proportionally to $n$, and unless the attention scale grows with $n$, the weights $A_{ij}$ approach $1/n$ and drive the system toward uniform mixing as in the Kuramoto model.  
This phenomenon amplifies the contractive dynamics described above and accelerates clustering and hence representation collapse.  
Motivated by this effect, several practical long-context systems including Qwen~\cite{bai2023qwen}, SSMax~\cite{nakanishi2025scalable}, and SWAN-GPT~\cite{puvvada2025swan} adopt a logarithmic attention scaling $\beta_n = \Theta(\log n)$.

To capture this dependence on the sequence length, we let the attention scale depend on $n$ through
\[
\beta_n=\gamma\log n,\qquad \gamma>0.
\]
With this choice, the attention weights may be written as
\[
A_{ij}=\frac{e^{\beta_n\langle x_i, x_j\rangle}}{\sum_{k=1}^n e^{\beta_n\langle x_i, x_k\rangle}}
=\frac{n^{\gamma \langle x_i, x_j\rangle}}{\sum_{k=1}^n n^{\gamma \langle x_i, x_k\rangle}}.
\]
In the equiangular model, we have $\langle x_i, x_j\rangle=\rho$ for $i \neq j$ and $\langle x_i,x_i\rangle=1$. Hence
\[
A_{ij}
=\frac{n^{\gamma \rho}}{n^\gamma + (n-1)n^{\gamma \rho}}
\sim
\begin{cases}
n^{-1}, & \gamma < \frac{1}{1-\rho},\\[0.4em]
n^{-\gamma(1-\rho)}, & \gamma > \frac{1}{1-\rho},
\end{cases}
\]
which already reveals two qualitatively different behaviors.  
For $\gamma<\frac{1}{1-\rho}$, the weights $A_{ij}$ are asymptotically uniform, so each token interacts with almost all others, and the layer behaves as an averaging operator.  
Conversely, for $\gamma>\frac{1}{1-\rho}$, the diagonal term dominates and the attention mechanism becomes effectively suppressed.  
The boundary $\gamma=\frac{1}{1-\rho}$ corresponds to a critical regime where attention concentrates on a sublinear yet nontrivial set of neighbors, preserving enough structure to propagate information without collapsing the tokens. This intuition may be summarized in the following result that shows contraction of the output directions of the attention layer (with Pre-LN) defined as:
$$
\mathrm{ATT}(x_i) = \sum_{j=1}^n x_j A_{ij}=r_i \cdot \theta_i, \quad r_i>0, \ \theta_i \in \Sph^{d-1}\,.
$$

\begin{theorem}[\cite{chen2025critical}]
\label{thm:long-context-phase-transition}
Assume that the inputs to the attention layer are equiangular: $\langle x_i, x_j\rangle=\rho$ if $i\neq j$ for some $\rho\in(0,1)$, $\|x_i\|=1$, and attention scale $\beta_n=\gamma\log n$. Then, the output directions $\theta_1,\ldots,\theta_n$ of a single attention layer satisfy for any $i \neq j$, 
\[
\lim_{n \to \infty}\langle \theta_i,\theta_j\rangle
=
\begin{cases}
1, & \gamma<\frac{1}{1-\rho},\\[0.4em]
\frac{4\rho}{1+3\rho}, & \gamma=\frac{1}{1-\rho},\\[0.6em]
\rho, & \gamma>\frac{1}{1-\rho}.
\end{cases}
\]
\end{theorem}

In particular, a single attention block already exhibits a phase transition:
\emph{uniform contraction} for subcritical $\gamma$, \emph{critical sparse mixing} at $\gamma=\frac{1}{1-\rho}$, and an \emph{identity-like regime} for supercritical $\gamma$.  
Since repeated layers amplify contraction multiplicatively, the one-step behavior fully determines long-time clustering.  
Thus the equiangular model provides a clean analytical description of how the logarithmic scaling $\beta_n\sim\log n$ stabilizes long-context attention by maintaining content-adaptive sparsity while avoiding collapse. Note also that a critical scaling of order $\beta \sim \log n$ also appears in perturbations of the exact equiangular model, in particular allowing dimension to be of order $\log n$ rather than $n$.

\section{Noisy Transformers.}
\label{sec:noisy}

Introducing noise into the attention dynamics leads to the following stochastic differential equation (SDE) on the sphere:
\[
    \ud X_i(t)
    = \proj_{X_i(t)}\!\left(\frac{1}{n} \sum_{j=1}^n e^{\beta \langle X_i(t), X_j(t)\rangle}X_j(t)\, \ud t \right)
    + \sqrt{2\kappa^{-1}}\, \ud W_i(t),
\]
where $\kappa>0$ controls the relative strength of the stochastic and drift terms, and $W_1,\dots,W_n$ are independent Brownian motions on $\mathbb{S}^{d-1}$. In the limit $\kappa \to \infty$, one recovers the deterministic dynamics~\eqref{eq:USA}. 

When $n$ is large and the initial conditions are i.i.d.\ with law $\mu_0$, the empirical distribution of the system converges to the solution of the McKean--Vlasov SDE
\begin{equation}
    \label{eq:McKV}
    \ud X(t)
    = \proj_{X(t)}\!\left(\int e^{\beta \langle X(t), y\rangle}\, y\mu_t(\ud y)\right)\ud t
    + \sqrt{2\kappa^{-1}}\, \ud W(t),
\end{equation}
where $\mu_t$ denotes the law of $X_t$. The corresponding evolution of $\mu_t$ satisfies the Fokker--Planck equation
\begin{equation}
    \label{eq:Fokker}
    \partial_t \mu_t
    + \kappa^{-1} \Delta \mu_t
    = \nabla \cdot \left( \mu_t\int e^{\beta\langle \cdot , y\rangle} y\, \mu_t(\ud y)\right),
\end{equation}
which reduces to~\eqref{eq:continuity} as $\kappa \to \infty$.

The bifurcation structure of stationary solutions to this \emph{noisy Transformer} model was first analyzed in~\cite{shalova2024solutions}, extending earlier work in~\cite{carrillo2020long}. For $d=2$, these results were sharpened in~\cite{banerjee}. As in the deterministic setting, the noisy Transformer dynamics interpolate with the classical \emph{noisy Kuramoto} model, obtained by taking $\beta=0$. This special case is known under various names—including the \emph{mean-field plane rotator} and \emph{XY-spin} models—and its stationary and dynamical properties are well understood: the uniform distribution is the unique stationary solution for $\kappa \le 2$, a pitchfork bifurcation occurs at $\kappa=2$, and a unique (up to rotation) nontrivial branch exists for $\kappa>2$~\cite{bertini2010dynamical}. A significant body of work, culminating in the uniform-in-time propagation-of-chaos result of~\cite{DelTse25}, now gives a remarkably complete picture of the noisy Kuramoto model.

By contrast, a quantitative description of the noisy Transformer dynamics remains largely open. Phenomena such as metastability suggest that the strong uniform-in-time results available for the noisy Kuramoto model may not hold in this richer setting; see, e.g.,~\cite{GarPapYan17, hairer}. Moreover, many natural variations, such as introducing common noise, exploring different geometries, modifying the interaction kernel, or considering anisotropic or multiplicative noise, lead to further mathematical challenges. Understanding these variants, and charting the full phase diagram of noisy attention dynamics, presents a wide landscape of open problems.

\section*{Acknowledgments.}
I am most grateful to  Borjan Geshkovski, Shi Chen, Zhengjiang Lin, and Krishna Balasubramanian for many insightful comments on early versions of this text.

\bibliographystyle{alphaabbr}
\bibliography{biblioICM}

\end{document}